\documentclass[runningheads]{llncs}
\usepackage{graphicx}
\usepackage{wrapfig}
\usepackage{multirow}
\usepackage[table,xcdraw]{xcolor}
\usepackage[normalem]{ulem}
\usepackage{booktabs}
\usepackage{hyperref}
\usepackage{amsmath}
\usepackage{marvosym}
\newcommand{\Envelope}{\textsuperscript{(\scalebox{1.0}{\Letter})}}
\usepackage[ruled,linesnumbered]{algorithm2e}

\useunder{\uline}{\ul}{}
\begin{document}
\title{Gaussian Mixture Modeling for Event-Aware Visual Allocation in Long Video Understanding}
\titlerunning{GMM-EVA}

%
%


\author{Yifan Lu\inst{1,2,3} \and
Ziqi Zhang\inst{1,2} \and
Chunfeng Yuan\inst{1,2,3}\Envelope\and \\
Jun Gao\inst{4} \and
Bing Li\inst{1,2,3} \and
Weiming Hu\inst{1,2,3,5}
}

\authorrunning{Y. Lu et al.}

\institute{
Beijing Key Laboratory of Super Intelligent Security of Multi-Modal Information, CASIA \and
State Key Laboratory of Multimodal Artificial Intelligence Systems, CASIA \and
School of Artificial Intelligence, University of Chinese Academy of Sciences \and
Hello Group \and
School of Information Science and Technology, ShanghaiTech University 
\email{luyifan2021@ia.ac.cn, cfyuan@nlpr.ia.ac.cn}
}


\maketitle           

\begin{abstract}
Keyframe selection has emerged as an effective paradigm to mitigate the prohibitive visual token overhead of Large Vision-Language Models (LVLMs) in long video understanding.
Existing selection methods often treat video frames as isolated units and allocate visual budgets equally, thereby overlooking high-level semantic structures and introducing substantial intra-event redundancy.
To address these limitations, we propose GMM-EVA (\textbf{G}aussian \textbf{M}ixture \textbf{M}odeling for \textbf{E}vent-Aware \textbf{V}isual \textbf{A}llocation), a training-free, plug-and-play keyframe allocation method for long video understanding.
GMM-EVA leverages a Gaussian Mixture Model fitted via the EM algorithm to transform noisy frame-wise relevance scores into a structured representation of latent semantic events. 
An event-aware allocation strategy is then applied to preserve one primary high-resolution keyframe per event for high-fidelity detail, while utilizing lower-resolution secondary keyframes to maintain temporal context and optimize token budgets.
Extensive experiments on multiple long video benchmarks demonstrate that GMM-EVA significantly outperforms uniform sampling and achieves comparable or superior performance to state-of-the-art selection methods while consuming only approximately half of their visual token budget.
Furthermore, GMM-EVA generalizes robustly across diverse relevance measures and downstream LVLMs, highlighting its effectiveness, efficiency, and broad applicability.

\end{abstract}

\section{Introduction}

Large Vision-Language Models (LVLMs) have achieved remarkable progress in video understanding by leveraging powerful cross-modal alignment and temporal modeling capabilities\cite{bai2025qwen3,chen2024internvl25,lin2024video,zhang2024llava}.
However, these models typically rely on uniform temporal sampling, which becomes increasingly problematic for long-duration videos: the growing number of visual tokens escalates computational cost and often exceeds the constrained context windows of LVLMs.
This forces a suboptimal trade-off between temporal and spatial resolution, inevitably leading to significant information loss and hindering the practical deployment of LVLMs for long video understanding.

\begin{wrapfigure}[13]{r}{0.4\textwidth} 
\vspace{-18pt}
   \includegraphics[width=0.4\textwidth]{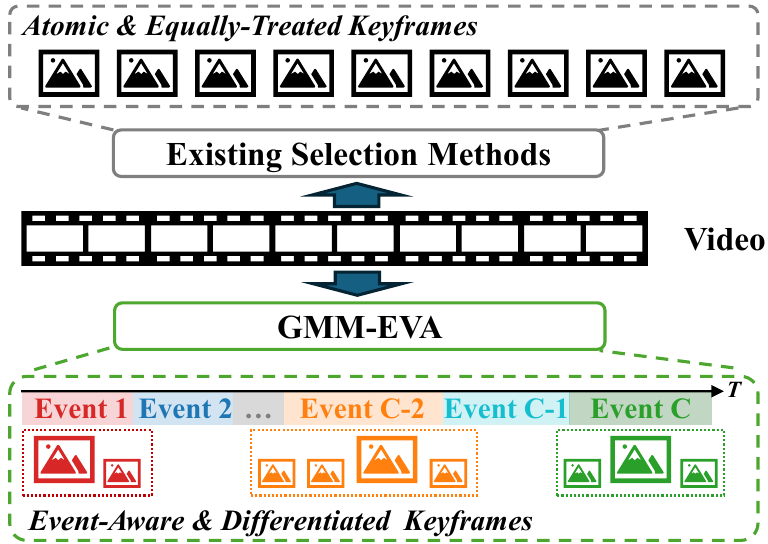}
    \caption{Comparison of the existing keyframe selection methods and our GMM-EVA.} \label{fig:1} 
  \centering
\end{wrapfigure}

Keyframe selection offers an effective paradigm to balance efficiency and accuracy by distilling a concise, task-relevant subset of frames from a long video, thereby minimizing visual token overhead while preserving essential information for downstream LVLMs.
Recent studies identify informative keyframes by either developing trainable selection modules 
\cite{tan2026msjoe,tang2026tspo,wang2025videoitg,yao2025generative,yu2025frame} 
or designing heuristic training-free policies \cite{liu2025bolt,ma2026gift,tang2025aks,ye2025tstar,zhu2025focus} based on frame-wise query-relevance scores computed by off-the-shelf cross-modal embedding models (e.g., CLIP~\cite{radford2021learning}).
However, as illustrated in Fig.\ref{fig:1}, existing methods treat video frames as isolated units, neglecting intrinsic temporal coherence and higher-level semantic structures. 
Moreover, they adopt an equal allocation strategy that assigns identical token budgets to all selected keyframes regardless of their semantic significance, wasting computational resources on temporally redundant or marginally informative frames.


To address these issues, we propose GMM-EVA (\textbf{G}aussian \textbf{M}ixture \textbf{M}odeling for \textbf{E}vent-Aware \textbf{V}isual \textbf{A}llocation), a training-free, plug-and-play method that explicitly models event-level structure to adaptively distribute the visual token budget (Fig.\ref{fig:1}).
The core of our approach involves parameterizing the temporal importance distribution via an event-centric Gaussian Mixture Model (GMM), where each component characterizes a latent semantic event. 
This formulation is motivated by the fact that events are high-level semantic concepts crucial for video understanding tasks \cite{du2024towards,lavee2009understanding,ma2024ea}, and the inherent smoothness and multi-peak nature of GMM make it well-suited for representing the continuous temporal span of multiple key events \cite{wang2024weakly,zheng2022cpl}.
Concretely, frame-wise query-relevance scores serve as discrete observations of latent temporal importance. By applying the Expectation-Maximization (EM) algorithm \cite{dempster1977maximum}, GMM-EVA transforms these noisy observations into a structured, continuous representation where latent events are explicitly defined by individual Gaussian components. 
By operating purely on scalar relevance scores, GMM-EVA decouples the event discovery process from any specific embedding model or downstream LVLM, enabling seamless integration across diverse model combinations.

Building on this event-level representation, we further propose an event-aware visual allocation strategy.
Each keyframe is assigned to its most probable latent event via GMM posterior probabilities, and within each event the most salient frame is designated as the primary keyframe while the rest serve as secondary keyframes.
By allocating high-resolution budgets to primary keyframes as semantic anchors and low-resolution budgets to secondary keyframes for temporal context, this design prunes intra-event redundancy and optimizes the token budget without compromising performance.

To summarize, our contributions are three-fold: 
\textbf{1)} We employ a Gaussian Mixture Model fitted via the EM algorithm to transform noisy frame-wise relevance scores into a structured representation of latent semantic events, enabling explicit temporal structuring of long videos.
\textbf{2)} Building on the discovered events, we design an event-aware allocation strategy that concentrates high-resolution budgets on primary keyframes as semantic anchors while using low-resolution secondary keyframes to maintain temporal context, effectively pruning intra-event redundancy.
\textbf{3)} Extensive experiments on multiple long video benchmarks show that GMM-EVA achieves competitive or superior performance while consuming approximately half the visual token budget, with robust generalization across diverse embedding backbones and downstream LVLMs.

\section{Related Works}

\subsection{Keyframe Selection for Long Video Understanding}
Keyframe selection aims to identify a condensed, task-relevant subset of frames or clips as input for downstream Large Vision-Language Models (LVLMs). 
Existing methodologies are generally categorized into training-based and training-free approaches.

\textbf{Training-based} approaches
\cite{tan2026msjoe,tang2026tspo,wang2025videoitg,yao2025generative,yu2025frame} 
typically train specialized modules for frame-wise scoring or index prediction. 
Methods such as VideoITG \cite{wang2025videoitg} and GenS \cite{yao2025generative} achieve high precision by employing LVLMs as selectors, yet suffer from significant inference latency and are constrained by the selector's own context window.
Alternatives like TSPO \cite{tang2026tspo} and MSJoE \cite{tan2026msjoe} reduce computational cost with lightweight trainable selectors, but their reliance on large-scale annotated datasets remains a major scalability bottleneck.

\textbf{Training-free} methods~\cite{liu2025bolt,ma2026gift,sun2025f2c,tang2025aks,ye2025tstar,zhang2025qframe,zhu2025focus}
leverage off-the-shelf cross-modal embedding models, such as CLIP \cite{radford2021learning}, to measure query-frame relevance without additional training. 
Beyond basic TopK selection \cite{liu2025bolt}, recent heuristics like AKS \cite{tang2025aks} and BOLT \cite{liu2025bolt} refine keyframe composition but still treat sampled frames as atomic and equal units. 
F2C \cite{sun2025f2c} extends selected anchors to surrounding clips, and Q-Frame \cite{zhang2025qframe} applies global resolution adjustments, yet neither captures higher-level temporal semantics.
In contrast, GMM-EVA explicitly models event-level structure via generative modeling, transforming noisy frame-wise scores into a structured event representation and enabling adaptive, differentiated allocation within each event.

\subsection{Gaussian Temporal Modeling in Video Understanding}
Gaussian distributions are widely adopted as temporal priors in video understanding, owing to their inherent smoothness and ability to model continuous temporal spans, particularly in temporal grounding tasks \cite{li2023d3g,long2019gaussian,zheng2022cpl}. 
For instance, GTAN \cite{long2019gaussian} and CPL \cite{zheng2022cpl} utilize learnable Gaussian masks as temporal weights to aggregate visual features.
To capture the multi-peak nature of video content \cite{zhao2025multi}, recent methods \cite{kim2024gaussian,wang2024weakly} extend single Gaussian masks to Gaussian Mixture Models (GMMs), enabling the representation of multiple query-relevant events.
Departing from these learnable approaches, GMM-EVA estimates GMM parameters directly from frame-wise relevance scores via the EM algorithm \cite{dempster1977maximum}, enabling structured event discovery without supervised training or gradient-based optimization.
\section{Methodology}
\begin{figure}[t]
\centering
   \includegraphics[width=0.92\textwidth]{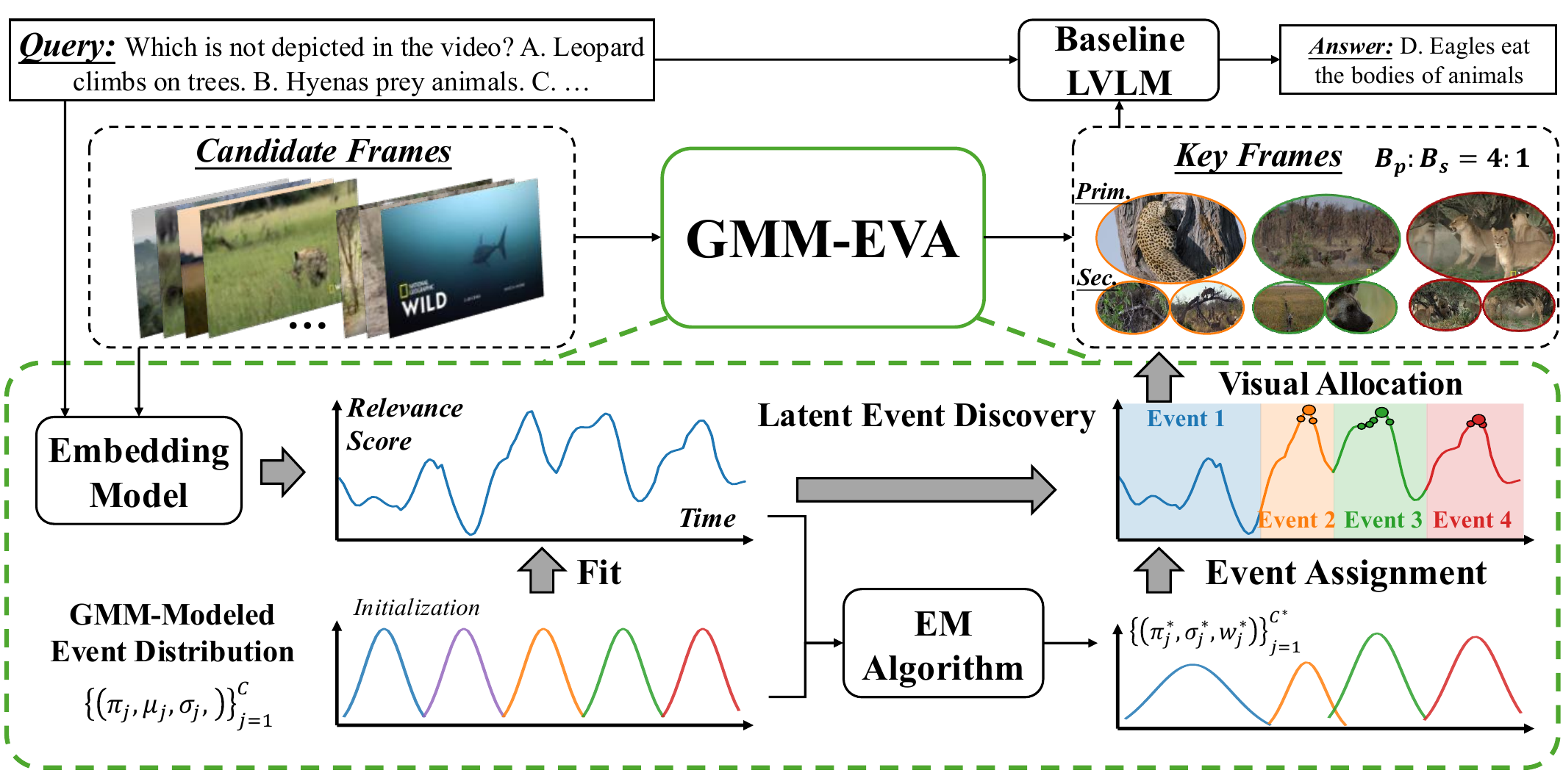}
    \caption{Overview of the proposed GMM-EVA pipeline. 1) A cross-modal embedding model computes frame-wise query-relevance scores. 2) A Gaussian Mixture Model (GMM), fitted via the EM algorithm, decomposes the score distribution into multiple latent semantic events. 3) An event-aware allocation strategy differentiates visual budgets, designating one high-resolution primary keyframe and multiple low-resolution secondary keyframes per event. The keyframes in the two sets are then fed into a downstream LVLM for answer generation.} \label{fig:overview} 
\end{figure}

Given a video with $T$ candidate frames $\mathcal{F} = \{f_t\}_{t=1}^T$ and a textual query $Q$, GMM-EVA selects a compact subset of keyframes as the visual input for a downstream LVLM.
As illustrated in Fig.~\ref{fig:overview} and Algorithm~\ref{alg:gmm_eva}, the pipeline consists of three stages.
First, a cross-modal embedding model computes frame-wise query-relevance scores as observations of temporal importance (\S\ref{sec:3.1}).
Then, a Gaussian Mixture Model is fitted to these observations via the EM algorithm, decomposing the score distribution into latent semantic events (\S\ref{sec:3.2}).
Finally, an event-aware allocation strategy assigns each keyframe to its most probable event and differentiates between a high-resolution primary keyframe and lower-resolution secondary keyframes within each event, yielding two subsets $\mathcal{F}_{prim}$ and $\mathcal{F}_{sec}$ (\S\ref{sec:3.3}).

\begin{algorithm}[t]
\caption{GMM-EVA: Gaussian Event-aware Visual Allocation}
\label{alg:gmm_eva}
\KwIn{Candidate Frames $\mathcal{F}=\{f_t\}_{t=1}^T$, Textual Query $Q$, Initial Component Interval $\Delta t_c$, Merging Threshold $\epsilon_m$, Target Allocation Number $K$.}
\KwOut{Primary Keyframe Set $\mathcal{F}_{prim}$, Secondary Keyframe Set $\mathcal{F}_{sec}$.}
\tcp{Stage 1: Query-Relevance Modeling}
$S = \{s_t\}_{t=1}^T \leftarrow \text{sim}(\text{CLIP}_V(\mathcal{F}), \text{CLIP}_T(Q))$ \tcp*{Eq.\ref{eq:1}}
$S' = \text{Normalize}(S)$ \tcp*{Eq.\ref{eq:2}}

\tcp{Stage 2: Latent Event Discovery via GMM}
$\Theta = \{\pi_j, \mu_j, \sigma_j\}_{j=1}^{C} \leftarrow \text{UniformInitialize}(T, \Delta t_c )$ \tcp*{Eq.\ref{eq:4}}
\While{not converged}{
    \text{E-Step: Update responsibilities} $\{\gamma_{t,j}\}$ \text{ using current} $\Theta$ \tcp*{Eq.\ref{eq:5}}
    \text{M-Step: Re-estimate} $\Theta \text{ based on } S' \text{ and } \{\gamma_{t,j}\} $ \tcp*{Eq.\ref{eq:6}}
    $\Theta \leftarrow \text{MergeComponents}(\Theta)$  \tcp*{Eq.\ref{eq:7}}
}
$\Theta^* = \{\pi_j^*, \mu_j^*, \sigma_j^*\}_{j=1}^{C^*} \leftarrow \Theta$

\tcp{Stage 3: Event-aware Visual Allocation}
Sort $\mathcal{F}$ into $\{f_{(1)}, f_{(2)}, \dots, f_{(T)}\}$ s.t. $s_{(1)} \ge s_{(2)} \ge \dots \ge s_{(T)}$ \;

$\mathcal{F}_{prim} = \emptyset, \mathcal{F}_{sec} = \emptyset, \mathcal{E}_{Top\text{-}K}= \emptyset$ \;

\For{$k = 1$ \KwTo $K$}{

    $j^*\leftarrow \text{EventAssign}((k),\Theta^*)$ \tcp*{Eq.\ref{eq:8}} 
    \eIf{$j^*$ not in $\mathcal{E}_{Top\text{-}K}$}{
        $\mathcal{F}_{prim} = \mathcal{F}_{prim} \cup \{f_{(k)}\}$, $\mathcal{E}_{Top\text{-}K}=\mathcal{E}_{Top\text{-}K} \cup \{j^*\}$ \tcp*{New event}
    }{
        $\mathcal{F}_{sec} = \mathcal{F}_{sec} \cup \{f_{(k)}\}$ \tcp*{Event already has a primary keyframe}
    }
}
\Return $\mathcal{F}_{prim}, \mathcal{F}_{sec}$
\end{algorithm}


\subsection{Query-Relevance Modeling}
\label{sec:3.1}
We first quantify the semantic relevance of each candidate frame $f_t$ to the textual query $Q$. Specifically, GMM-EVA employs CLIP \cite{radford2021learning}, a lightweight pretrained cross-modal embedding model, whose visual and text encoders are denoted by $\text{CLIP}_V(\cdot)$ and $\text{CLIP}_T(\cdot)$, respectively. The query-relevance score $s_t$ is computed as the cosine similarity between the frame and query embeddings:
\begin{equation}
    s_t = \text{sim}\bigl(\text{CLIP}_V(f_t),\ \text{CLIP}_T(Q)\bigr),
    \label{eq:1}
\end{equation}
where $\text{sim}(\cdot,\cdot)$ denotes the cosine similarity.
To interpret the scores set $S = \{s_t\}_{t=1}^T$ as a valid probability distribution for subsequent GMM fitting, we apply a linear normalization:
\begin{equation}
    s_t' =  \frac{s_t-\min(S)}{\sum_{i=1}^T (s_i-\min(S))}.
    \label{eq:2}
\end{equation}
The resulting set $S' = \{s_t'\}_{t=1}^T$ serves as a discrete, empirical observation of how temporal importance is distributed across the video, providing the evidence required for latent event modeling in the next stage.

\subsection{Latent Event Discovery via GMM}
\label{sec:3.2}
To move beyond isolated frame-level analysis and capture the coherent temporal structure of videos, GMM-EVA employs a Gaussian Mixture Model (GMM) to parametrically model the query-conditioned importance distribution.
Specifically, the temporal importance density $p(\tau)$ is formulated as a mixture of $C$ Gaussian components, each representing a latent semantic event:
\begin{equation}
p(\tau) = \sum_{j=1}^{C} \pi_j \mathcal{N}(\tau | \mu_j, \sigma_j^2), ~s.t. \sum_{j=1}^{C} \pi_j=1, 
\label{eq:3}
\end{equation}
where $\tau \in [1,T]$ denotes the continuous temporal coordinate and $\mathcal{N}$ denotes the Gaussian probability density function. Since candidate frames are uniformly sampled at a fixed rate, the temporal coordinate of frame $f_t$ is identified with its index, i.e., $\tau = t$.
For each component, the mixing coefficient $\pi_j$, mean $\mu_j$, and standard deviation $\sigma_j$ encode the event's cumulative significance, temporal center, and duration, respectively.

The parameters $\Theta = \{\pi_j, \mu_j, \sigma_j\}_{j=1}^{C} $ of the GMM are estimated by applying the Expectation-Maximization (EM) algorithm to the discrete observations $S'$. 
GMM-EVA first adopts a uniform initialization strategy with a temporal stride $\Delta t_c$. The initial number of components is set to $C = \lfloor \frac{T}{\Delta t_c} \rfloor$, with the $j$-th component initialized as:
\begin{equation}
\pi_j=\frac{1}{C},\quad\mu_j = \frac{jT}{C+1},\quad\sigma_j=\frac{T}{2C},\quad\text{for}\quad j=1,2,\dots,C.
\label{eq:4}
\end{equation}
The EM algorithm then alternates between the E-step and M-step until convergence. 
The E-step computes the responsibility $\gamma_{t,j}$, i.e., the posterior probability that frame $f_t$ belongs to the $j$-th latent event:
\begin{equation}
\gamma_{t,j} = \frac{\pi_j \mathcal{N}(t | \mu_j, \sigma_j^2)}{\sum_{i=1}^{C} \pi_i \mathcal{N}(t | \mu_i, \sigma_i^2)}.
\label{eq:5}
\end{equation}
The M-step updates the parameters to maximize the weighted log-likelihood. 
Unlike standard EM, GMM-EVA incorporates $s_t'$ as an importance weight for each frame.
Let $N_j = \sum_{t=1}^T s_t' \gamma_{t,j}$ denote the effective count of the $j$-th component; the parameters are updated as follows:
\begin{equation}
\begin{aligned}
\pi_j = \frac{N_j}{\sum_{k=1}^{C} N_k} = N_j,\quad
\mu_j &= \frac{1}{N_j} \sum_{t=1}^T s_t' \gamma_{t,j} t,\quad \sigma_j = \sqrt{\frac{1}{N_j} \sum_{t=1}^T s_t' \gamma_{t,j} (t - \mu_j)^2}\\
\text{for}\quad t&=1,2,...,T,~j=1,2,...,C,
\end{aligned}
\label{eq:6}
\end{equation}
Furthermore, to refine the event structure, we incorporate a component merging scheme \cite{long2019gaussian}. Two components $\{\pi_{(1)}, \mu_{(1)}, \sigma_{(1)}\}$ and $\{\pi_{(2)}, \mu_{(2)}, \sigma_{(2)}\}$ are merged if the Intersection over Union (IoU) of their 1-$\sigma$ intervals exceeds a threshold $\epsilon_m$. The merged parameters are derived as:
\begin{equation}
\begin{aligned}
\pi_{new} &= \pi_{(1)} + \pi_{(2)},\quad 
\mu_{new} = \frac{\mu_{(1)}\pi_{(1)} + \mu_{(2)}\pi_{(2)}}{\pi_{new}},\\
\sigma_{new} &= \sqrt{\frac{\pi_{(1)}\sigma_{(1)}^2 + \pi_{(2)}\sigma_{(2)}^2}{\pi_{(1)} + \pi_{(2)}} + \frac{\pi_{(1)}\pi_{(2)}(\mu_{(1)} - \mu_{(2)})^2}{(\pi_{(1)} + \pi_{(2)})^2}}.
\end{aligned}
\label{eq:7}
\end{equation}

As outlined in Alg. \ref{alg:gmm_eva}, the EM iteration continues until the increment of the weighted log-likelihood $\mathcal{L}(\Theta) = \sum_{t=1}^T s_t' \ln \left( \sum_{j=1}^{C} \pi_j \mathcal{N}(t | \mu_j, \sigma_j^2) \right)$ falls below a threshold $\epsilon_L$, or the iteration count reaches a maximum limit $I_{max}$.
The converged parameter set $\Theta^* = \{\pi_j^*, \mu_j^*, \sigma_j^*\}_{j=1}^{C^*}$ constitutes a structured representation of the video's event-level temporal layout, where each Gaussian component probabilistically characterizes a latent semantic event, forming the foundation for the subsequent visual budget allocation.

\subsection{Event-Aware Visual Allocation}
\label{sec:3.3}
With the latent event structure established via $\Theta^*$, GMM-EVA implements a differentiated strategy to allocate visual budgets by categorizing keyframes into primary and secondary roles based on their event membership.
We first sort the candidate frames in descending order of their relevance scores $S$ and select the top-$K$ allocation targets. 
For each top-$K$ frame $f_{(k)}$, we compute its posterior probability of belonging to the $j$-th latent event using the converged parameters $\Theta^*$ via Eq.~\ref{eq:5}, denoted as $\gamma^*_{(k),j}$.
GMM-EVA performs a hard assignment by associating the frame with its most probable event $j^*$, formulated as:
\begin{equation}
    j^* = \arg\max_{j \in {1, \dots, C^*}} \gamma^*_{(k),j}
    \label{eq:8}
\end{equation}
The union of events assigned to the top-$K$ frames constitutes the active event set $\mathcal{E}_{Top\text{-}K}$.

To mitigate intra-event redundancy, we differentiate frames within each active event $j \in \mathcal{E}_{Top\text{-}K}$: the highest-scoring frame is designated as the primary keyframe and collected into $\mathcal{F}_{prim}$, serving as the semantic anchor of that event, while all remaining frames form the secondary set $\mathcal{F}_{sec}$, providing auxiliary temporal context at a lower visual budget.
Since the top-$K$ frames are traversed in descending relevance order, this categorization is realized by checking whether each frame's assigned event has been seen before (Algorithm~\ref{alg:gmm_eva}).

When fed into downstream LVLMs, primary keyframes in $\mathcal{F}_{prim}$ are allocated a higher visual budget $B_p$ (e.g., high resolution), while secondary keyframes in $\mathcal{F}_{sec}$ receive a lower budget $B_s$ (e.g., low resolution). 
This differentiated approach concentrates computational resources on unique semantic anchors while maintaining temporal continuity through low-cost secondary frames, effectively optimizing the token budget without compromising understanding performance.
Notably, GMM-EVA can be flexibly scaled by incorporating events beyond $\mathcal{E}_{Top\text{-}K}$, as further analyzed in Sec.~\ref{sec:scale}.

\section{Experiments}

\subsection{Experimental Setups}
\subsubsection{Implementation Details}
In our experiments, videos are sampled at 1 FPS to obtain candidate frames. 
We employ CLIP~\cite{radford2021learning} as the default cross-modal embedding model, with additional experiments conducted using SigLIP~\cite{zhai2023sigmoid}, LanguageBind-Video~\cite{zhu2024languagebind}, and Qwen3-VL-Embedding-8B~\cite{li2026qwen3}. 
For GMM parameters, the temporal stride $\Delta t_c$ is set to 30, the merging threshold $\epsilon_m$ to 0.7, the log-likelihood convergence threshold $\epsilon_L$ to $10^{-5}$, and the maximum iteration count $I_{max}$ to 1000. 
The top-$K{=}32$ target frames by relevance score are selected for event-aware allocation, with $B_p{=}256$ tokens per primary keyframe and $B_s{=}64$ tokens per secondary keyframe. 
We adopt Qwen2.5-VL-7B as the default downstream LVLM, with generalization experiments further conducted using Qwen2-VL-7B~\cite{wang2024qwen2}, Qwen3-VL-8B~\cite{bai2025qwen3}, and InternVL-2.5-8B~\cite{chen2024internvl25}. 
Unless otherwise specified, all experiments use CLIP and Qwen2.5-VL-7B.

\subsubsection{Evaluation}
We evaluate GMM-EVA on three representative long-form video understanding benchmarks: \textbf{LongVideoBench}~\cite{wu2024longvideobench}, \textbf{LVBench}~\cite{wang2025lvbench}, and \textbf{Video-MME}~\cite{fu2025videomme}. 
For LongVideoBench, we additionally construct a \emph{multi} split comprising question types that involve multiple events (\emph{E3E}, \emph{O3O}, \emph{SSS}, \emph{SOS}, and \emph{SAA}, per the official taxonomy) to specifically assess temporal structure modeling.
For VideoMME, we additionally report results on the \emph{long} split (average duration 2,466.7s).
We use multiple-choice accuracy as the primary metric. Computational efficiency is measured by the relative token budget ratio $\mathcal{R}_B$, defined as the percentage of consumed tokens relative to a baseline that processes all $K{=}32$ frames at $B_p{=}256$ tokens each.

\begin{table}[t]
\centering
\caption{Performance comparison of keyframe selection methods with Qwen2.5-VL-7B as the downstream LVLM. For GMM-EVA, $\mathcal{R}_B$ values are listed in the order of LongVideoBench/LVBench/VideoMME. The best and second-best results are highlighted in \textbf{bold} and {\ul underline}, respectively.}\label{tab:1}
\resizebox{0.8\columnwidth}{!}{
\begin{tabular}{lccccccc}
\toprule
& & \multicolumn{2}{c}{LongVideoBench} & & \multicolumn{2}{c}{VideoMME} \\
\multirow{-2}{*}{Method}           & \multirow{-2}{*}{$\mathcal{R}_B$(\%)}         & overall          & multi            & \multirow{-2}{*}{LVBench} & overall        & long  
& \multirow{-2}{*}{Average} \\
\midrule
\multicolumn{7}{l}{\emph{\scriptsize Baseline methods}} \\
Uniform        & \multirow{2}{*}{100} & 57.4       & 54.4       & 36.9       & 60.9       & 50.1    & 53.4       \\
Top-K            &  & {\ul 61.3} & {\ul 57.0} & {\ul 49.6} & 61.3       & 50.8  & 58.0       \\
\midrule
\multicolumn{7}{l}{\emph{\scriptsize Training-free methods}} \\
AKS \cite{tang2025aks} & \multirow{3}{*}{100}  & 59.2       & 53.5       & 46.4       & 61.9       & 52.2    & 56.9       \\
Q-Frame \cite{zhang2025qframe}       &  & 58.7       & -          & -          & 62.6       & 53.1    & -       \\
BOLT \cite{liu2025bolt}         &  & 60.4       & 55.1       & 43.2       & 63.0       & 53.7    & 56.9       \\ \midrule
\multicolumn{7}{l}{\emph{\scriptsize Training-based methods}} \\
VideoITG \cite{wang2025videoitg}      & \multirow{2}{*}{100} & 59.8       & 55.7       & 47.2       & 63.2       & \textbf{55.0}  & 57.9 \\
MSJoE \cite{tan2026msjoe}          &  & 60.1       & -          & 46.4       & \textbf{64.3} & 54.1 & {\ul 58.3}      \\ \midrule
\rowcolor[HTML]{DDDDDD} GMM-EVA w/CLIP & {\scriptsize 44.9/52.7/44.8}   & 61.2       & {\ul 57.0}      & {\ul 49.6} & 61.1       & 50.9 & 57.9       \\
\rowcolor[HTML]{DDDDDD} GMM-EVA w/Qwen & {\scriptsize 44.2/49.9/43.8} & \textbf{62.7}    & \textbf{58.9}   & \textbf{53.3}    & {\ul 63.8}  & {\ul 54.2} & \textbf{60.6} \\ 
\bottomrule
\end{tabular}
}
\end{table}

\subsection{Main Results}
We compare GMM-EVA with several representative keyframe selection methods in Tab.\ref{tab:1}. All competing methods consume the full 100\% budget (32 keyframes $\times$ 256 tokens each) for downstream LVLM visual input.

With the CLIP backbone, GMM-EVA uses only \textbf{44.9\%}/\textbf{52.7\%}/\textbf{44.8\%} of the base budget on LongVideoBench/LVBench/VideoMME, yet achieves competitive performance. Compared with \textbf{baseline} methods, GMM-EVA with CLIP outperforms uniform sampling by a large margin across all benchmarks, and matches the Top-K baseline in accuracy while consuming approximately half the token budget.
When compared with \textbf{training-free} methods, although GMM-EVA with CLIP does not surpass all competitors on every benchmark under its reduced budget, it achieves the best score on the \emph{multi} split of LongVideoBench, underscoring the advantage of explicit event modeling for questions involving multiple events.
For \textbf{training-based} methods, their specialized selection modules are trained to precisely measure frame-query relevance, thus outperforming GMM-EVA when paired with the lightweight CLIP encoder. Nevertheless, when equipped with the more powerful Qwen3-VL-Embedding-8B, GMM-EVA surpasses VideoITG/MSJoE by +2.7/+2.3 on average.
These results validate the effectiveness and efficiency of GMM-EVA for long video understanding.

\begin{table}[t]
\centering
\begin{minipage}[]{0.53\textwidth}
\centering
\caption{Generalization across embedding backbones on LongVideoBench.}\label{tab:2}
\resizebox{\columnwidth}{!}{
\begin{tabular}{llcc}
\toprule
Embedding Model &
  Method &
  $\mathcal{R}_B$(\%) &
\begin{tabular}[c]{@{}l@{}}LongVideo\\ \quad Bench\end{tabular}  \\
\midrule
- & Uniform & 100   & 57.4            \\
\midrule
\multirow{3}{*}{SigLIP}                                                           
& Top-K & 50.0   & 59.6 \\
& Top-K & 100   & 60.4 \\
& \cellcolor[HTML]{DDDDDD}GMM-EVA & \cellcolor[HTML]{DDDDDD}44.3 & \cellcolor[HTML]{DDDDDD}61.0  \\
\midrule
\multirow{3}{*}{LanguageBind-Video}    
& Top-K & 50.0   & 59.2  \\
& Top-K & 100.0   & 60.5  \\
& \cellcolor[HTML]{DDDDDD}GMM-EVA& \cellcolor[HTML]{DDDDDD} 41.4  & \cellcolor[HTML]{DDDDDD}60.6 \\
\midrule
\multirow{3}{*}{Qwen3-VL-Emb-8B} 
& Top-K & 50.0   & 61.4 \\
& Top-K & 100   & 62.3 \\
& \cellcolor[HTML]{DDDDDD}GMM-EVA& \cellcolor[HTML]{DDDDDD}44.2  & \cellcolor[HTML]{DDDDDD}62.7\\
\bottomrule
\end{tabular}
}
\end{minipage}
\hfill
\begin{minipage}[]{0.45\textwidth}
\centering
\caption{Compatibility with downstream LVLMs on LongVideoBench.}\label{tab:3}
\resizebox{\columnwidth}{!}{
\begin{tabular}{llcc}
\toprule
\begin{tabular}[c]{@{}l@{}}Downstream\\ LVLM\end{tabular} &
  Method &
  $\mathcal{R}_B$(\%) &
\begin{tabular}[c]{@{}l@{}}LongVideo\\ \quad Bench\end{tabular}  \\
\midrule
\multirow{6}{*}{Qwen2-VL-7B}     
& Uniform & 100 & 57.0 \\
& Top-K & 50.0     & 58.6 \\
& Top-K & 100     & 59.1 \\
& GenS \cite{yao2025generative} & 84.4      & 58.7    \\
& TSPO \cite{tang2026tspo} & 82.0     & 58.6    \\
& \cellcolor[HTML]{DDDDDD}GMM-EVA & \cellcolor[HTML]{DDDDDD}44.9  & \cellcolor[HTML]{DDDDDD}58.7 \\
\midrule
\multirow{4}{*}{Qwen3-VL-8B}     
& Uniform & 100 & 62.5    \\
& Top-K & 50.0     & 62.2  \\
& Top-K & 100     & 65.7  \\
& \cellcolor[HTML]{DDDDDD}GMM-EVA & \cellcolor[HTML]{DDDDDD}44.9    & \cellcolor[HTML]{DDDDDD}65.2     \\
\midrule
\multirow{4}{*}{InternVL-2.5-8B} 
& Uniform & 100 & 60.7 \\
& Top-K & 50.0     & 63.1\\
& Top-K & 100     & 63.2 \\ 
& \cellcolor[HTML]{DDDDDD}GMM-EVA & \cellcolor[HTML]{DDDDDD} 44.9  & \cellcolor[HTML]{DDDDDD}65.3  \\
\bottomrule
\end{tabular}
}
\end{minipage}
\end{table}

\subsection{Analysis and Ablation Studies}
\subsubsection{Generalization across Embedding Backbones}
As shown in Tab.\ref{tab:2}, GMM-EVA yields consistent gains of \textbf{+3.6}/\textbf{+3.2}/\textbf{+5.3} over uniform sampling on LongVideoBench with SigLIP/LanguageBind-Video/Qwen3-VL-Embedding-8B, while matching or exceeding the full-budget Top-K baseline ($\mathcal{R}_B{=}100\%$) using roughly half the tokens and surpassing Top-K at comparable budgets ($\mathcal{R}_B{=}50\%$).
Among the evaluated backbones, Qwen3-VL-Embedding-8B yields the best results, as its stronger cross-modal alignment provides higher-quality importance observations for GMM estimation.
These results demonstrate that GMM-EVA generalizes seamlessly across diverse relevance sources, and its performance ceiling can be further elevated by integrating more advanced embedding models.

\subsubsection{Compatibility with Downstream LVLMs}
As shown in Tab.\ref{tab:3}, GMM-EVA improves over uniform sampling by \textbf{+1.7}/\textbf{+2.7}/\textbf{+4.6} on LongVideoBench when paired with Qwen2-VL-7B/Qwen3-VL-8B/InternVL-2.5-8B.
Across three downstream LVLMs, GMM-EVA with only \textbf{44.9\%} of the token budget consistently outperforms Top-K at 50\% budget and achieves comparable accuracy to Top-K at 100\%.
Notably, on Qwen2-VL-7B, GMM-EVA also matches two training-based methods, GenS and TSPO, without any training and with substantially fewer tokens.
The consistent efficacy across different model families (Qwen-VL and InternVL) and versions highlights the strong compatibility of our method.


\begin{figure}[t]
\centering
   \includegraphics[width=0.92\textwidth]{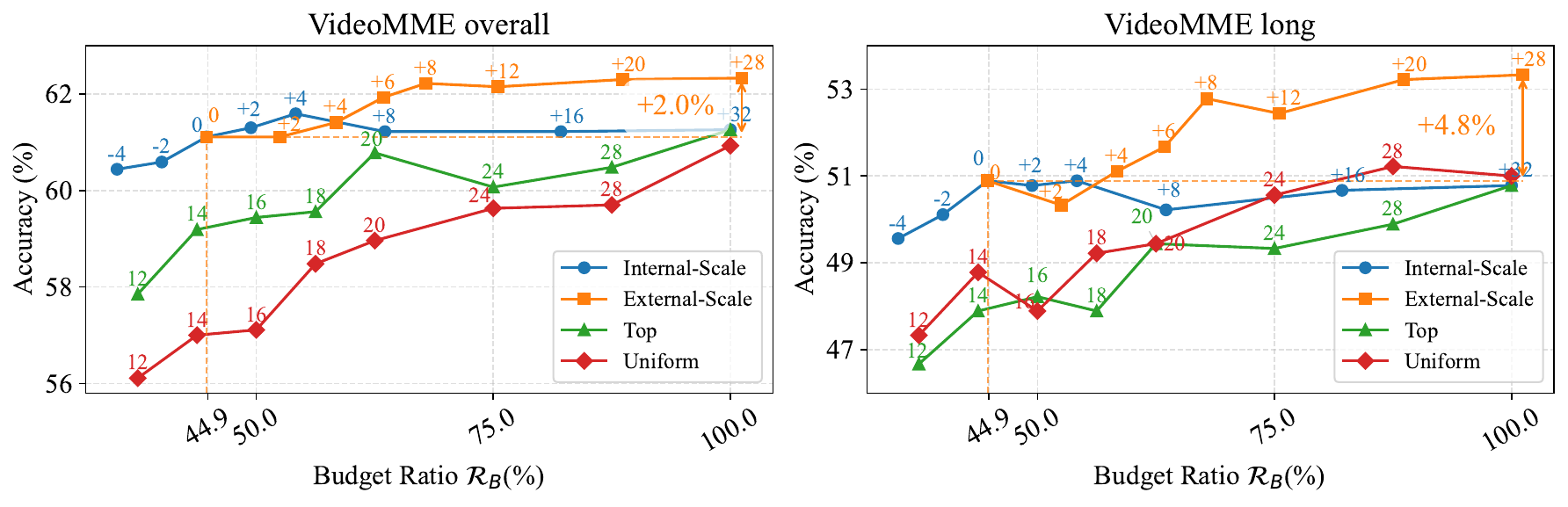}
    \caption{Scaling analysis on VideoMME. Numbers above data points of two scaling methods denote the incremental primary keyframes added; numbers above Uniform and Top-K denote total frame counts.} \label{fig:scale} 
\end{figure}

\subsubsection{Scaling Analysis}
\label{sec:scale}
We analyze the scalability of GMM-EVA under varying token budgets.
As discussed in Sec.\ref{sec:3.3}, GMM-EVA can be scaled by incorporating \textbf{external} latent events beyond the targeted $K$ keyframes. We empirically allocate 1 primary and 2 secondary keyframes per additional event.
We compare this with an \textbf{internal} variant that upgrades existing secondary keyframes to primary within the $K$ targeted frames, as well as \textbf{Uniform} and \textbf{Top-K} baselines with varying frame counts.

As shown in Fig.\ref{fig:scale}, the External strategy provides superior expansion capability over the Internal approach, and GMM-EVA consistently outperforms both baselines across all budget levels, demonstrating that our method enhances understanding performance even at equivalent budget scales.
Furthermore, when externally scaled with 28 additional events, GMM-EVA achieves a \textbf{4.8\%} improvement on the \emph{long} split, substantially exceeding the \textbf{2.0\%} gain on the full VideoMME set, suggesting that event-aware modeling is particularly advantageous for long-form videos.

\begin{table}[t]
\centering
\begin{minipage}[]{0.61\textwidth}
\centering
\caption{Analysis on primary keyframe allocation strategies.}\label{tab:4}
\resizebox{\columnwidth}{!}{
\begin{tabular}{lcccc}
\toprule
\begin{tabular}[c]{@{}l@{}}Allocation \\ Method\end{tabular} & 
\begin{tabular}[c]{@{}l@{}}LongVideo\\ \quad Bench\end{tabular} & LVBench & VideoMME & Average \\
\midrule
Top     & 60.4          & 44.4          & 59.4  & 55.5        \\
K-means & 60.6          & 44.2          & 59.3  & 55.4          \\
\rowcolor[HTML]{DDDDDD}  Event-Aware     & 61.2 & 49.6 & 61.1 & 57.9 \\
\bottomrule
\end{tabular}
}
\end{minipage}
\hfill
\begin{minipage}[]{0.37\textwidth}
\centering
\caption{Ablation on secondary keyframe token budget $B_s$.}\label{tab:5}
\resizebox{\columnwidth}{!}
{
\begin{tabular}{lccc}
\toprule
Method & $B_s$ & $\mathcal{R}_B$(\%) & VideoMME \\
\midrule
Top-K & 0 & 25.0     & 57.6  \\

\midrule
& 0                        & 26.2  & 59.4                                 \\
& 16                       & 30.9  & 60.4                                 \\
& 32                       & 35.5 & 60.2                                 \\
& \cellcolor[HTML]{DDDDDD}64 & \cellcolor[HTML]{DDDDDD}44.8 & \cellcolor[HTML]{DDDDDD} 61.1 \\
 & 128                     & 63.2  & 60.6     \\
\multirow{-6}{*}{GMM-EVA} & 256 & 100 & 61.3 \\
\bottomrule
\end{tabular}
}
\end{minipage}
\end{table} 

\subsubsection{Primary Keyframe Allocation}
We compare the \textbf{event-aware} allocation of GMM-EVA against two alternatives for designating primary keyframes:
1) \textbf{Top}, which directly assigns the highest-scoring frames to $\mathcal{F}_{prim}$;
2) \textbf{K-means}, which clusters keyframe timestamps and selects the highest-scoring frame per cluster.
For fair comparison, $|\mathcal{F}_{prim}|$ is set to the number of events $C^*$ adaptively determined by GMM-EVA for each video, providing the two baselines with an oracle event count.

As shown in Tab.\ref{tab:4}, the event-aware strategy achieves the best performance across all benchmarks, validating the effectiveness of GMM-based adaptive event modeling.
The Top method tends to select temporally adjacent frames, lacking diversity; K-means attempts temporal partitioning but cannot distinguish events with non-uniform densities.
Moreover, despite receiving the oracle event count from GMM-EVA, neither baseline can match our method, as they lack the ability to adaptively discover the optimal event structure and must rely on external guidance for this critical hyperparameter.

\subsubsection{Impact of Secondary Keyframe Budget}
We vary the per-frame token budget $B_s$ of secondary keyframes to assess their contribution.
As shown in Tab.\ref{tab:5}, $B_s{=}64$ yields the best accuracy; lower values incur notable information loss, while higher values bring only marginal gains at substantially greater cost.
Removing secondary keyframes entirely ($B_s{=}0$) causes a \textbf{1.7}-point drop compared to $B_s{=}64$, confirming the importance of the temporal context they provide.
Notably, even without secondary frames, GMM-EVA surpasses the Top-K baseline at comparable budgets by \textbf{1.8}, further validating the superiority of our event-aware selection in distilling higher-quality semantic keyframes.

\begin{figure}[t]
\centering
   \includegraphics[width=0.95\textwidth]{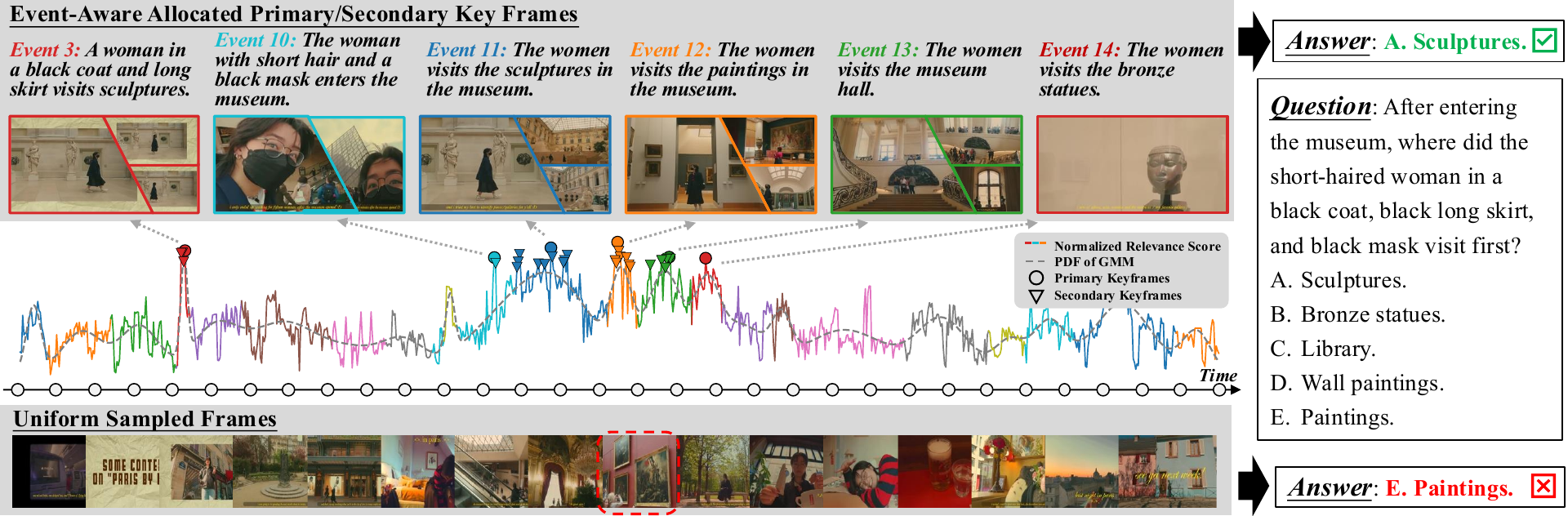}
    \caption{Qualitative comparison between GMM-EVA and uniform sampling. The solid curve shows normalized relevance scores with color-coded event segments; the dashed curve depicts the GMM-estimated importance density. Primary and secondary keyframes are marked with circles and triangles, respectively, colored by their event assignments.} \label{fig:vis} 
\end{figure}

\subsection{Qualitative Analysis}
Fig.\ref{fig:vis} presents a qualitative comparison. The dashed GMM density closely fits the empirical score curve, confirming that the model faithfully captures the underlying event structure.
GMM-EVA correctly identifies the key events relevant to the question (e.g., ``visit sculpture'', ``enter museum'') and allocates representative keyframes accordingly, producing a correct answer.
In contrast, uniform sampling misses critical temporal cues and only captures ``painting''-related content, leading to an incorrect answer.

\section{Conclusion}
In this paper, we revisited keyframe selection for long video understanding through the lens of event-level semantics, arguing that recognizing latent events offers a principled path toward more efficient visual budget utilization.
Guided by this insight, GMM-EVA recasts noisy relevance scores into a structured Gaussian mixture, upon which an event-aware allocation strategy distinguishes primary semantic anchors from secondary contextual frames.
Across three long-video benchmarks, GMM-EVA achieved competitive or superior accuracy with roughly half the conventional token budget, generalizing consistently across diverse embedding backbones and downstream LVLMs.
The gain became more pronounced on the long-duration split of VideoMME, suggesting event-level structuring as a promising direction for further scaling LVLMs to longer videos.

\subsubsection*{Acknowledgements}
This work was supported by Beijing Natural Science Foundation (L243015), Beijing Major Science and Technology Project under Contract no. Z251100008425008.




\bibliographystyle{splncs04}
\bibliography{ref}

@article{dempster1977maximum,
  title={Maximum likelihood from incomplete data via the EM algorithm},
  author={Dempster, Arthur P and Laird, Nan M and Rubin, Donald B},
  journal={Journal of the royal statistical society: series B (methodological)},
  volume={39},
  number={1},
  pages={1--22},
  year={1977},
  publisher={Wiley Online Library}
}

@inproceedings{ma2024ea,
  title={Ea-vtr: Event-aware video-text retrieval},
  author={Ma, Zongyang and Zhang, Ziqi and Chen, Yuxin and Qi, Zhongang and Yuan, Chunfeng and Li, Bing and Luo, Yingmin and Li, Xu and Qi, Xiaojuan and Shan, Ying and others},
  booktitle={European Conference on Computer Vision},
  pages={76--94},
  year={2024},
  organization={Springer}
}

@article{lavee2009understanding,
  title={Understanding video events: A survey of methods for automatic interpretation of semantic occurrences in video},
  author={Lavee, Gal and Rivlin, Ehud and Rudzsky, Michael},
  journal={IEEE Transactions on Systems, Man, and Cybernetics, Part C (Applications and Reviews)},
  volume={39},
  number={5},
  pages={489--504},
  year={2009},
  publisher={IEEE}
}

@article{du2024towards,
  title={Towards event-oriented long video understanding},
  author={Du, Yifan and Zhou, Kun and Huo, Yuqi and Li, Yifan and Zhao, Wayne Xin and Lu, Haoyu and Zhao, Zijia and Wang, Bingning and Chen, Weipeng and Wen, Ji-Rong},
  journal={arXiv preprint arXiv:2406.14129},
  year={2024}
}

@inproceedings{radford2021learning,
  title={Learning transferable visual models from natural language supervision},
  author={Radford, Alec and Kim, Jong Wook and Hallacy, Chris and Ramesh, Aditya and Goh, Gabriel and Agarwal, Sandhini and Sastry, Girish and Askell, Amanda and Mishkin, Pamela and Clark, Jack and others},
  booktitle={International conference on machine learning},
  pages={8748--8763},
  year={2021},
  organization={PmLR}
}

@inproceedings{zhu2024languagebind,
  title={LanguageBind: Extending Video-Language Pretraining to N-modality by Language-based Semantic Alignment},
  author={Zhu, Bin and Lin, Bin and Ning, Munan and Yan, Yang and Cui, Jiaxi and HongFa, WANG and Pang, Yatian and Jiang, Wenhao and Zhang, Junwu and Li, Zongwei and others},
  booktitle={The Twelfth International Conference on Learning Representations},
  year={2024}
}

@inproceedings{zhai2023sigmoid,
  title={Sigmoid loss for language image pre-training},
  author={Zhai, Xiaohua and Mustafa, Basil and Kolesnikov, Alexander and Beyer, Lucas},
  booktitle={Proceedings of the IEEE/CVF international conference on computer vision},
  pages={11975--11986},
  year={2023}
}

@article{li2026qwen3,
  title={Qwen3-VL-Embedding and Qwen3-VL-Reranker: A Unified Framework for State-of-the-Art Multimodal Retrieval and Ranking},
  author={Li, Mingxin and Zhang, Yanzhao and Long, Dingkun and Chen, Keqin and Song, Sibo and Bai, Shuai and Yang, Zhibo and Xie, Pengjun and Yang, An and Liu, Dayiheng and others},
  journal={arXiv preprint arXiv:2601.04720},
  year={2026}
}

@inproceedings{lin2024video,
  title={Video-llava: Learning united visual representation by alignment before projection},
  author={Lin, Bin and Ye, Yang and Zhu, Bin and Cui, Jiaxi and Ning, Munan and Jin, Peng and Yuan, Li},
  booktitle={Proceedings of the 2024 conference on empirical methods in natural language processing},
  pages={5971--5984},
  year={2024}
}

@article{zhang2024llava,
  title={Llava-video: Video instruction tuning with synthetic data},
  author={Zhang, Yuanhan and Wu, Jinming and Li, Wei and Li, Bo and Ma, Zejun and Liu, Ziwei and Li, Chunyuan},
  journal={arXiv preprint arXiv:2410.02713},
  year={2024}
}

@article{wang2024qwen2,
  title={Qwen2-vl: Enhancing vision-language model's perception of the world at any resolution},
  author={Wang, Peng and Bai, Shuai and Tan, Sinan and Wang, Shijie and Fan, Zhihao and Bai, Jinze and Chen, Keqin and Liu, Xuejing and Wang, Jialin and Ge, Wenbin and others},
  journal={arXiv preprint arXiv:2409.12191},
  year={2024}
}

@article{bai2025qwen3,
  title={Qwen3-vl technical report},
  author={Bai, Shuai and Cai, Yuxuan and Chen, Ruizhe and Chen, Keqin and Chen, Xionghui and Cheng, Zesen and Deng, Lianghao and Ding, Wei and Gao, Chang and Ge, Chunjiang and others},
  journal={arXiv preprint arXiv:2511.21631},
  year={2025}
}

@article{chen2024internvl25,
  title={Expanding performance boundaries of open-source multimodal models with model, data, and test-time scaling},
  author={Chen, Zhe and Wang, Weiyun and Cao, Yue and Liu, Yangzhou and Gao, Zhangwei and Cui, Erfei and Zhu, Jinguo and Ye, Shenglong and Tian, Hao and Liu, Zhaoyang and others},
  journal={arXiv preprint arXiv:2412.05271},
  year={2024}
}

@article{wu2024longvideobench,
  title={Longvideobench: A benchmark for long-context interleaved video-language understanding},
  author={Wu, Haoning and Li, Dongxu and Chen, Bei and Li, Junnan},
  journal={Advances in Neural Information Processing Systems},
  volume={37},
  pages={28828--28857},
  year={2024}
}

@inproceedings{wang2025lvbench,
  title={Lvbench: An extreme long video understanding benchmark},
  author={Wang, Weihan and He, Zehai and Hong, Wenyi and Cheng, Yean and Zhang, Xiaohan and Qi, Ji and Ding, Ming and Gu, Xiaotao and Huang, Shiyu and Xu, Bin and others},
  booktitle={Proceedings of the IEEE/CVF International Conference on Computer Vision},
  pages={22958--22967},
  year={2025}
}

@inproceedings{fu2025videomme,
  title={Video-mme: The first-ever comprehensive evaluation benchmark of multi-modal llms in video analysis},
  author={Fu, Chaoyou and Dai, Yuhan and Luo, Yongdong and Li, Lei and Ren, Shuhuai and Zhang, Renrui and Wang, Zihan and Zhou, Chenyu and Shen, Yunhang and Zhang, Mengdan and others},
  booktitle={Proceedings of the IEEE/CVF conference on computer vision and pattern recognition},
  pages={24108--24118},
  year={2025}
}

@article{sun2025f2c,
  title={From frames to clips: Training-free adaptive key clip selection for long-form video understanding},
  author={Sun, Guangyu and Singhal, Archit and Uzkent, Burak and Shah, Mubarak and Chen, Chen and Kessler, Garin},
  journal={arXiv preprint arXiv:2510.02262},
  year={2025}
}

@inproceedings{zhang2025qframe,
  title={Q-frame: Query-aware frame selection and multi-resolution adaptation for video-llms},
  author={Zhang, Shaojie and Yang, Jiahui and Yin, Jianqin and Luo, Zhenbo and Luan, Jian},
  booktitle={Proceedings of the IEEE/CVF International Conference on Computer Vision},
  pages={22056--22065},
  year={2025}
}

@inproceedings{ye2025tstar,
  title={Re-thinking temporal search for long-form video understanding},
  author={Ye, Jinhui and Wang, Zihan and Sun, Haosen and Chandrasegaran, Keshigeyan and Durante, Zane and Eyzaguirre, Cristobal and Bisk, Yonatan and Niebles, Juan Carlos and Adeli, Ehsan and Fei-Fei, Li and others},
  booktitle={Proceedings of the IEEE/CVF Conference on Computer Vision and Pattern Recognition},
  pages={8579--8591},
  year={2025}
}

@article{ma2026gift,
  title={GIFT: Global Irreplaceability Frame Targeting for Efficient Video Understanding},
  author={Ma, Junpeng and Zhou, Sashuai and Li, Guanghao and Gao, Xin and Cao, Yue and Zeng, Hengyu and Yan, Yuxiang and Wang, Zhibin and Song, Jun and Zheng, Bo and others},
  journal={arXiv preprint arXiv:2603.25072},
  year={2026}
}

@article{zhu2025focus,
  title={Focus: Efficient keyframe selection for long video understanding},
  author={Zhu, Zirui and Xu, Hailun and Luo, Yang and Liu, Yong and Sarkar, Kanchan and Yang, Zhenheng and You, Yang},
  journal={arXiv preprint arXiv:2510.27280},
  year={2025}
}

@article{wang2025videoitg,
  title={VideoITG: Multimodal Video Understanding with Instructed Temporal Grounding},
  author={Wang, Shihao and Chen, Guo and Huang, De-an and Li, Zhiqi and Li, Minghan and Li, Guilin and Alvarez, Jose M and Zhang, Lei and Yu, Zhiding},
  journal={arXiv preprint arXiv:2507.13353},
  year={2025}
}

@inproceedings{liu2025bolt,
  title={Bolt: Boost large vision-language model without training for long-form video understanding},
  author={Liu, Shuming and Zhao, Chen and Xu, Tianqi and Ghanem, Bernard},
  booktitle={Proceedings of the Computer Vision and Pattern Recognition Conference},
  pages={3318--3327},
  year={2025}
}

@inproceedings{tang2025aks,
  title={Adaptive keyframe sampling for long video understanding},
  author={Tang, Xi and Qiu, Jihao and Xie, Lingxi and Tian, Yunjie and Jiao, Jianbin and Ye, Qixiang},
  booktitle={Proceedings of the Computer Vision and Pattern Recognition Conference},
  pages={29118--29128},
  year={2025}
}

@inproceedings{yu2025frame,
  title={Frame-voyager: Learning to query frames for video large language models.(2025)},
  author={Yu, Sicheng and Jin, Chengkai and Wang, Huanyu and Chen, Zhenghao and Jin, Sheng and Zuo, Zhongrong and Xu, Xiaolei and Sun, Zhenbang and Zhang, Bingni and Wu, Jiawei and others},
  booktitle={Proceedings of the Thirteenth International Conference on Learning Representations, ICLR},
  pages={24--28},
  year={2025}
}

@article{tan2026msjoe,
  title={MSJoE: Jointly Evolving MLLM and Sampler for Efficient Long-Form Video Understanding},
  author={Tan, Wenhui and Yu, Xiaoyi and Li, Jiaze and Chen, Yijing and Ju, Jianzhong and Luo, Zhenbo and Song, Ruihua and Luan, Jian},
  journal={arXiv preprint arXiv:2602.22932},
  year={2026}
}

@inproceedings{tang2026tspo,
  title={Tspo: Temporal sampling policy optimization for long-form video language understanding},
  author={Tang, Canhui and Han, Zifan and Sun, Hongbo and Zhou, Sanping and Zhang, Xuchong and Wei, Xin and Yuan, Ye and Zhang, Huayu and Xu, Jinglin and Sun, Hao},
  booktitle={Proceedings of the AAAI Conference on Artificial Intelligence},
  volume={40},
  number={11},
  pages={9368--9376},
  year={2026}
}

@inproceedings{yao2025generative,
  title={Generative frame sampler for long video understanding},
  author={Yao, Linli and Wu, Haoning and Ouyang, Kun and Zhang, Yuanxing and Xiong, Caiming and Chen, Bei and Sun, Xu and Li, Junnan},
  booktitle={Findings of the Association for Computational Linguistics: ACL 2025},
  pages={17900--17917},
  year={2025}
}

@inproceedings{long2019gaussian,
  title={Gaussian temporal awareness networks for action localization},
  author={Long, Fuchen and Yao, Ting and Qiu, Zhaofan and Tian, Xinmei and Luo, Jiebo and Mei, Tao},
  booktitle={Proceedings of the IEEE/CVF conference on computer vision and pattern recognition},
  pages={344--353},
  year={2019}
}

@inproceedings{zheng2022cpl,
  title={Weakly supervised temporal sentence grounding with gaussian-based contrastive proposal learning},
  author={Zheng, Minghang and Huang, Yanjie and Chen, Qingchao and Peng, Yuxin and Liu, Yang},
  booktitle={Proceedings of the IEEE/CVF Conference on Computer Vision and Pattern Recognition},
  pages={15555--15564},
  year={2022}
}

@inproceedings{li2023d3g,
  title={D3g: Exploring gaussian prior for temporal sentence grounding with glance annotation},
  author={Li, Hanjun and Shu, Xiujun and He, Sunan and Qiao, Ruizhi and Wen, Wei and Guo, Taian and Gan, Bei and Sun, Xing},
  booktitle={Proceedings of the IEEE/CVF International Conference on Computer Vision},
  pages={13734--13746},
  year={2023}
}

@inproceedings{wang2024weakly,
  title={Weakly supervised gaussian contrastive grounding with large multimodal models for video question answering},
  author={Wang, Haibo and Lai, Chenghang and Sun, Yixuan and Ge, Weifeng},
  booktitle={Proceedings of the 32nd ACM International Conference on Multimedia},
  pages={5289--5298},
  year={2024}
}

@inproceedings{kim2024gaussian,
  title={Gaussian mixture proposals with pull-push learning scheme to capture diverse events for weakly supervised temporal video grounding},
  author={Kim, Sunoh and Cho, Jungchan and Yu, Joonsang and Yoo, YoungJoon and Choi, Jin Young},
  booktitle={Proceedings of the AAAI Conference on Artificial Intelligence},
  volume={38},
  number={3},
  pages={2795--2803},
  year={2024}
}

@inproceedings{zhao2025multi,
  title={Multi-Granularity Distribution Modeling for Video Watch Time Prediction via Exponential-Gaussian Mixture Network},
  author={Zhao, Xu and Ma, Ruibo and Chen, Jiaqi and Zhao, Weiqi and Yang, Ping and Hu, Yao},
  booktitle={Proceedings of the Nineteenth ACM Conference on Recommender Systems},
  pages={309--318},
  year={2025}
}
\end{document}